\documentclass[lettersize,journal]{IEEEtran}
\usepackage{amsmath,amsfonts}
\usepackage{algorithmic}
\usepackage{array}
\usepackage{booktabs}
\usepackage[caption=false,font=normalsize,labelfont=sf,textfont=sf]{subfig}
\usepackage{textcomp}
\usepackage{stfloats}
\usepackage{url}
\usepackage{verbatim}
\usepackage{graphicx}
\usepackage{caption}
\def\BibTeX{{\rm B\kern-.05em{\sc i\kern-.025em b}\kern-.08em
    T\kern-.1667em\lower.7ex\hbox{E}\kern-.125emX}}
\usepackage{balance}
\usepackage{makecell}
\usepackage{float} 

\newfloat{teaser}{tbp}{lop} 
\floatname{teaser}{Teaser}  

\usepackage{xurl}
\begin{document}
\title{Geometry and Perception Guided Gaussians for Multiview-consistent 3D Generation from a Single Image}
\author{Pufan Li*, Bi'an Du*,~\IEEEmembership{Student Member, IEEE}, Wei Hu,~\IEEEmembership{Senior~Member, ~IEEE}
\thanks{
* Equal contribution. Pufan Li, Bi'an Du and Wei Hu are with Wangxuan Institute of Computer Technology, Peking University, No. 128, Zhongguancun North Street, Beijing, China (e-mail: lipufan@pku.edu.cn, pkudba@stu.pku.edu.cn, forhuwei@pku.edu.cn).

Corresponding author: Wei Hu (forhuwei@pku.edu.cn).
}
}

\markboth{Journal of \LaTeX\ Class Files,~Vol.~18, No.~9, September~2020}%
{How to Use the IEEEtran \LaTeX \ Templates}

\maketitle

\begin{figure*}[t!]
  \vspace{-0.2in}
  \includegraphics[width=\textwidth]{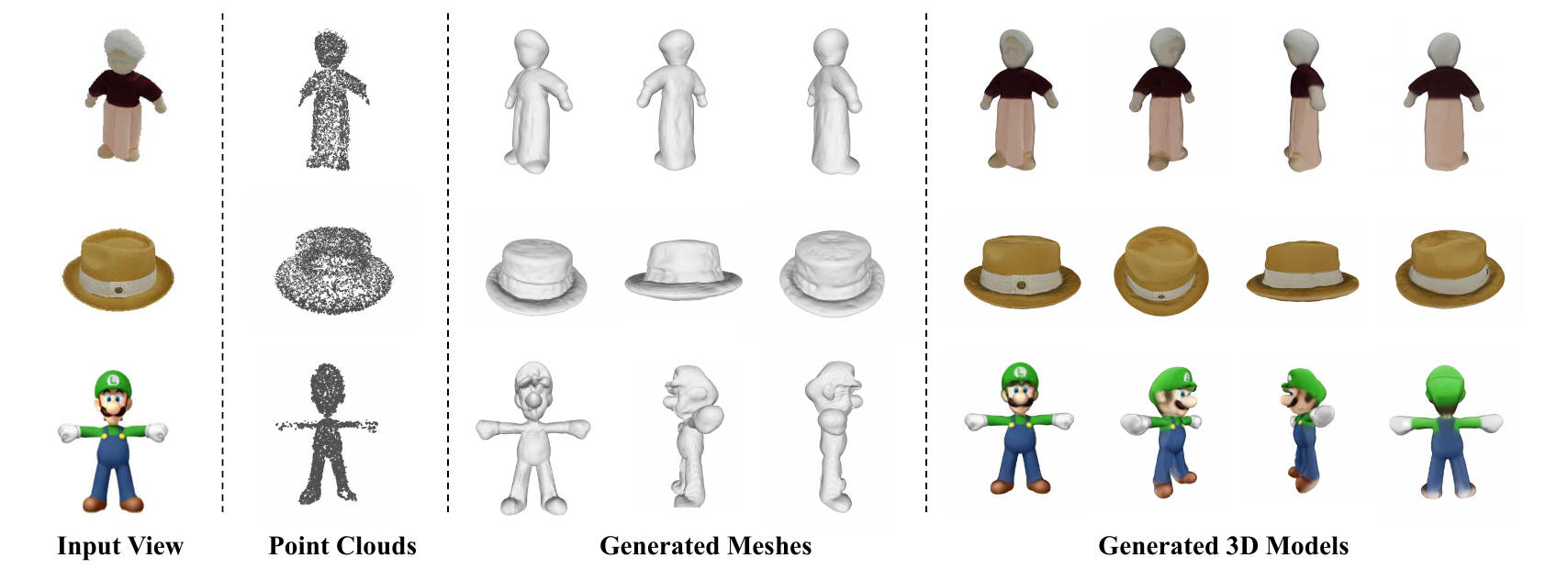}
  
  \caption{Our method is able to generate multiview-consistent 3D objects from a single-view input image. For each input image, our method generates point clouds, meshes and multiview images from any views without additional training.
  }
  \label{fig:teaser}
  \vspace{-0.05in}
\end{figure*}

\begin{abstract}
Generating realistic 3D objects from single-view images requires natural appearance, 3D consistency, and the ability to capture multiple plausible interpretations of unseen regions. Existing approaches often rely on fine-tuning pretrained 2D diffusion models or directly generating 3D information through fast network inference or 3D Gaussian Splatting, but their results generally suffer from poor multiview consistency and lack geometric detail. 
To tackle these issues, 
we present a novel method that seamlessly integrates geometry and perception information without requiring additional model training to reconstruct detailed 3D objects from a single image. 
Specifically, we incorporate geometry and perception priors to initialize the Gaussian branches and guide their parameter optimization.
The geometry prior captures the rough 3D shapes, while the perception prior utilizes the 2D pretrained diffusion model to enhance multiview information.
Subsequently, we introduce a stable Score Distillation Sampling for fine-grained prior distillation to ensure effective knowledge transfer. 
The model is further enhanced by a reprojection-based strategy that enforces depth consistency.
Experimental results show that we outperform existing methods on novel view synthesis and 3D reconstruction, demonstrating robust and consistent 3D object generation.
\end{abstract}



\vspace{-0.1in}
\section{Introduction}

Single-view 3D generation is a basic multi-modality task in multimedia, which aims to predict the full 3D geometry ({\it e.g.}, a mesh, point cloud) of an object or a scene from an input 2D image \cite{tang2023dreamgaussian, hu2024mvd, liu2023one, long2024wonder3d, qian2023magic123, melas2023realfusion,  liu2024seif, gao2022dasi, nie2023cpg3d}.
It requires that the resulting 3D content not only appears natural and consistent from all viewpoints, but also accounts for multiple plausible interpretations of unseen regions. Humans possess a remarkable ability to envision a complete 3D structure from a single image, even when certain parts of the object remain hidden. 
This gives credits to prior knowledge of familiar objects, so that the human brain infers and represents missing regions to form a coherent 3D concept.
In the fields of computer vision and graphics, substantial progress has been made in deriving implicit 3D representations from images \cite{xie2022neural,mildenhall2021nerf,wang2021neus}. However, explicitly reconstructing 3D objects directly from 2D images remains a formidable challenge.


Recent studies on single-view 3D generation can be broadly classified into two main categories: 3D native approaches and optimization-based 2D lifting approaches. 
The former \cite{wang2023rodin,jun2023shap,gupta20233dgen,du2024generative,muller2023diffrf,nichol2022point} relies on large, manually annotated datasets and significant computational resources to train large-scale neural networks, thereby enabling efficient, 3D-consistent outputs at inference time. However, these methods still struggle to produce realistic details.
Meanwhile, 2D diffusion models \cite{rombach2022high,ho2020denoising} have shown remarkable capabilities in image generation, opening new possibilities for 3D tasks. 
SyncDreamer \cite{liu2023syncdreamer} and MVD-Fusion \cite{hu2024mvd} extend 2D diffusion models to capture the joint probability distribution of multiview images, achieving satisfying novel view synthesis results.
On the other hand, diffusion models act as a bridge between 2D images and 3D objects. A prime example is Dreamfusion \cite{poole2022dreamfusion}, which introduces Score Distillation Sampling (SDS) to distill 3D geometry and appearance from a pretrained 2D diffusion model. This SDS technique has become central to many recent 2D lifting methods, where it is often used to optimize Neural Radiance Fields (NeRF) \cite{mildenhall2021nerf}, mitigating inconsistencies and ambiguities while modeling rich 3D content. 
Unlike NeRF-based approaches which require hours of optimization, 3D Gaussian Splatting \cite{kerbl20233d} achieves efficient reconstruction with significantly lower memory and time consumption.
DreamGaussian \cite{tang2023dreamgaussian} brings 3D Gaussian Splatting into the generative realm by using SDS to inject 2D prior knowledge into the 3D Gaussians, thus boosting generation efficiency while maintaining quality. However, its outputs still suffer from issues like blurry back views and oversaturated textures.

In this work, we propose a novel framework that seamlessly integrates geometry and perception priors to generate 3D objects from a single view, aiming to directly generating 3D consistent content while avoiding training or fine-turning additional neural networks at high computational and time expense.
Our method adopts 3D Gaussian Splatting as the 3D representation and enhances it with geometry and perception prior knowledge, merely involving optimizing the parameters of the Gaussians. 
We initialize three different Gaussian branches from the geometry prior, perception prior and Gaussian noise, aiming to integrate the advantages of different initializations. 
By leveraging the perception prior to augment the input image, we infuse the Gaussians with richer multiview information beyond that of the single input image. The introduction of the additional multiview information effectively reduces multiview inconsistency.
Compared with previous Gaussian-based single-view 3D generation methods \cite{tang2023dreamgaussian}, our approach leverages on broader prior knowledge to optimize the Gaussians.

Further, to improve the geometric details, we expand the SDS to a more stable version through directional constraint on its bias component for fine-grained prior distillation, eliminating blurry and oversaturated textures to some extent.
We also propose a reprojection-based method to exploit the discrepancy between the projected positions and the actual observed pixel locations in the rendered images from different Gaussian branches, further enhancing the consistency of generated results, as illustrated in Fig.~\ref{fig:teaser}.
In summary, our contributions are:

\begin{itemize}
  \item We introduce a novel pipeline for single-view 3D object generation by integrating geometry and perception prior knowledge into Gaussian Splatting. 
  
  \item We improve the stability of the SDS by constraining the direction of the bias component,
  which facilitates effective knowledge from pretrained diffusion models. We also propose a reprojection loss to enforce multiview consistency during the generation process. 

  
  \item The experimental results on the Google Scanned Object dataset \cite{downs2022google} demonstrate the superiority of our method compared to competing approaches. 
\end{itemize}

%



%

\vspace{-0.15in}
\section{Related Work}

\subsection{Generative 3D Representation}
A high-quality 3D representation is crucial for most 3D-related tasks.
Classic representations, including point clouds \cite{du2021self,fan2017point}, voxels \cite{choy20163d,maturana2015voxnet} and meshes \cite{wang2018pixel2mesh,wen2019pixel2mesh++}, offer detailed descriptions of geometry and appearance. However, they lack flexibility in the underlying topology and are difficult to optimize during the training process.

Neural Radiance Fields (NeRF) \cite{mildenhall2021nerf} and its variants \cite{li2023neuralangelo,barron2022mip,lin2023magic3d,poole2022dreamfusion} use neural networks to implicitly model 3D scenes, yielding high-quality reconstructions and realistic novel‐view images from sparse 2D inputs at the cost of heavy computation and large data requirements. In contrast, 3D Gaussian splatting \cite{kerbl20233d} represents scenes with continuous Gaussian distributions, offering both superior reconstruction quality and greater efficiency. DreamGaussian \cite{tang2023dreamgaussian} is the first work to leverage 3D Gaussian splatting for generation, achieving rapid processing and competitive visual fidelity.

\vspace{-0.15in}
\subsection{Diffusion Models for 3D Generation}
The diffusion model \cite{ho2020denoising,rombach2022high} has revolutionized 2D image generation, inspiring several attempts to extend it to 3D tasks \cite{jun2023shap,nichol2022point,du2024multi,luo2021diffusion}. However, the scarcity of 3D data makes direct training of diffusion models difficult, and even approaches that leverage only 2D images yield generation quality and generalization far below their 2D counterparts.

Many works, including DreamFusion \cite{poole2022dreamfusion}, SJC \cite{wang2023score}, RealFusion \cite{melas2023realfusion}, AugGS \cite{du2024auggs}, NeuralLift-360 \cite{xu2023neurallift} and DreamGaussian \cite{tang2023dreamgaussian}, leverage high-quality 2D diffusion models to bridge 2D and 3D data and thus enhance 3D inference without directly training a 3D diffusion model. DreamFusion and SJC formulate a score-distillation objective that uses a pretrained text-to-image diffusion model as a prior to generate 3D objects from text and to optimize neural radiance fields; RealFusion adapts a 2D diffusion model with text inversion to achieve full 360° reconstruction from a single image; NeuralLift-360 integrates a depth-aware NeRF with denoising diffusion models for scene generation; and DreamGaussian is the first to apply SDS to optimize 3D Gaussian Splatting. Together, these results demonstrate the effectiveness of distilling knowledge from 2D diffusion models for 3D tasks.

Some methods fine-tune 2D diffusion models to improve 3D performance. SparseFusion proposes a perspective diffusion model using an epipolar feature transformer and distillation for more accurate sparse-view reconstruction \cite{zhou2023sparsefusion}; Zero-1-to-3 adapts a large pretrained image diffusion model on 3D datasets to generate novel views from a single input, yielding better generalization \cite{liu2023zero}; and MVD-Fusion along with SyncDreamer employ a 3D-aware feature attention mechanism to synchronize intermediate states across multiview reverse diffusion steps, resulting in more consistent multiview images \cite{hu2024mvd,liu2023syncdreamer}.

\vspace{-0.15in}
\subsection{Single-view 3D Generation}
Single-view 3D generation seeks to produce 3D content from a single reference image. Recent methods reframe this as conditional novel-view synthesis \cite{liu2023zero,zhou2023sparsefusion,melas2023realfusion}, and others extend 2D diffusion models to multiview image generation by adapting large pretrained models for generalizable view synthesis \cite{chan2023generative,deng2023nerdi,gu2023nerfdiff,lei2022generative}. However, exclusive reliance on image-based synthesis paradigms fails to adequately resolve the challenges inherent in authentic 3D content generation.

Single-view 3D generation must yield fully detailed and interactive 3D content rather than only multiview images.
A complete 3D model is essential for applications in virtual reality, augmented reality and 3D printing, all of which require accurate depth, texture and geometry. 
Accordingly, the goal is to produce high-fidelity 3D models that capture fine-grained details, preserve structural consistency and support seamless manipulation and rendering from any viewpoint. 
Achieving this objective demands both improved consistency across views and innovations in 3D reconstruction techniques.
We argue that direct point cloud/voxel diffusion models and Gaussian splatting approaches are inherently superior for single-view 3D generation.


\vspace{-0.15in}
\section{Method}

\begin{figure*}[htbp]
  \centering
  \includegraphics[width=\textwidth]{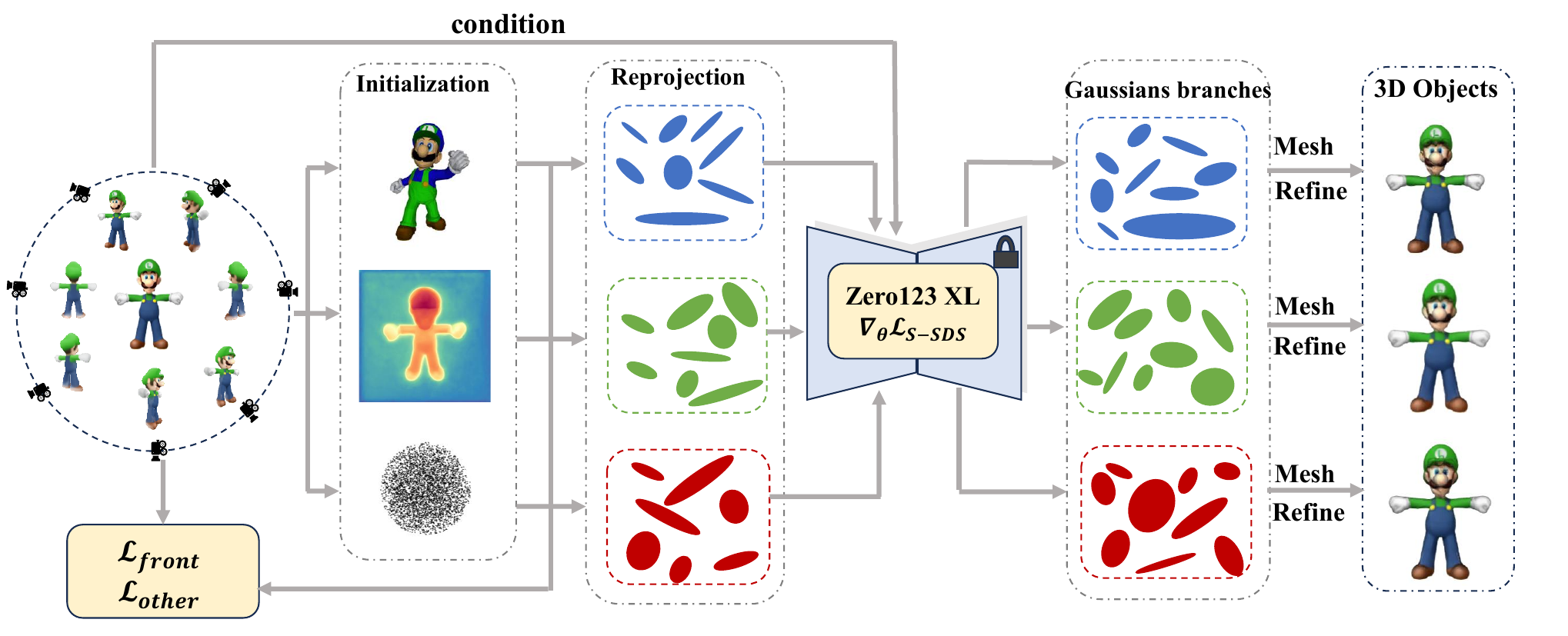}
  \vspace{-0.2in}
  \caption{The architecture of the proposed method. We use the geometry prior, perception prior and Gaussian noise to initialize three different Gaussian branches. At each training step, each Gaussian branch is supervised by the input image and augmented images generated by the perception prior.} 
  \label{fig:framework}
  \vspace{-0.1in}
\end{figure*}
Given a single image as input, our goal is to accurately generate the corresponding 3D object positioned at the origin.
We propose a novel framework for high-quality single-view 3D generation, with 3D Gaussian Splatting as our 3D representation, as shown in Fig.~\ref{fig:framework}.  
Our framework consists of four steps. 
1) We initialize three Gaussian branches with the geometry prior, perception prior and Gaussian noise. Each Gaussian branch is supervised by the input image and the augmented images generated by the perception prior. 
2) We then propose a reprojection loss to self-supervise the alignment of their corresponding image observations, enhancing the consistency of generated results.
3) To guarantee effective knowledge distillation, we propose a modified version of the standard SDS framework.
4) Finally, we adopt a mesh refinement module \cite{tang2023dreamgaussian}  to obtain a textured mesh.

In this section, we first provide the preliminaries. Then we introduce the consistency enhancement framework for single view 3D generation that combines geometry and perception priors simultaneously, aiming to fully use prior knowledge and further improve generation consistency.

\vspace{-0.1in}
\subsection{Preliminaries}
\label{sec:Preliminaries}
\subsubsection{3D Gaussian Splatting} 
3D Gaussian Splatting \cite{kerbl20233d} is an advanced technique in computer graphics and vision to efficiently represent and render complex 3D scenes. It enables fast and high-quality rendering through differentiable splatting operations. The method models 3D geometry and appearance using a set of anisotropic 3D Gaussian distributions. Each Gaussian is characterized by a set of parameters: a mean position $\mathbf{x} \in \mathbb{R}^3$, a scaling factor $\mathbf{s} \in \mathbb{R}^3$, a rotation quaternion $\mathbf{q} \in \mathbb{R}^4$ representing covariance, and attributes for color $\mathbf{c} \in \mathbb{R}^3$ and opacity $\alpha \in \mathbb{R}^3$. These parameters are collectively denoted as $\Theta$, where the parameters of the $i$-th Gaussian are given by $\Theta_i = \{\mathbf{x}_i, \mathbf{s}_i, \mathbf{q}_i, \alpha_i, \mathbf{c}_i\}$.

To determine the color of each pixel $\mathbf{p}$ in the image, a neural point-based rendering approach is employed. The rendering process proceeds as follows:
\begin{equation}
  \mathbf{C}(\mathbf{p})=\sum_{i\in\mathcal{N}}\mathbf{c}_i\alpha_i\prod_{j=1}^{i-1}(1-\alpha_i),
  \label{eq:gs_render}
\end{equation}
where $\mathcal{N}$ denotes the related Gaussians.

\subsubsection{From Diffusion Model to SDS Loss}

Diffusion models \cite{ho2020denoising} excel in generating high-fidelity samples and have demonstrated great success in applications including image synthesis, super-resolution, and inpainting. The diffusion framework consists of a forward diffusion process and a corresponding reverse (denoising) process.

In the forward process, the initial data $\mathbf{x}_0$ drawn from the data distribution $p(\mathbf{x})$ are progressively corrupted by introducing incremental noise in multiple time steps. 
Mathematically, this forward diffusion is modeled as a Markov chain:
\begin{equation}
    q(\mathbf{x}_t \mid \mathbf{x}_{t-1}) = \mathcal{N}(\mathbf{x}_t; \sqrt{1 - \beta_t}\, \mathbf{x}_{t-1}, \beta_t \mathbf{I}),
\end{equation}
where $\mathbf{x}_t$ denotes the noisy data at timestep $t$, and $\beta_t$ controls the noise schedule.

In the reverse process, the model learns to progressively remove noise from the corrupted data by reversing the forward diffusion. 
Mathematically, this reverse diffusion is also formulated as a Markov chain:
\begin{equation}
    p_{\theta}(\mathbf{x}_{t-1} \mid \mathbf{x}_t) = \mathcal{N}(\mathbf{x}_{t-1}; \mu_{\theta}(\mathbf{x}_t, t), \Sigma_{\theta}(\mathbf{x}_t, t)),
\end{equation}
where $\mu_{\theta}$ and $\Sigma_{\theta}$ represent parameters learned by the neural network.

Distilling knowledge from 2D diffusion models to optimize 3D representations has emerged as a promising approach for 3D generation tasks.
DreamFusion \cite{poole2022dreamfusion} introduces Score Distillation Sampling (SDS), an optimization method that leverages 2D diffusion priors for text-to-3D generation. Given a differentiable 3D model $f(\theta)$ capable of rendering images where $\theta$ represents the parameters of the 3D model, SDS provides gradient guidance toward high-probability image regions conditioned on input $\mathbf{y}$. The SDS loss gradient is defined as:
\begin{equation}
    \nabla_{\theta}\mathcal{L}_{\text{SDS}} = 
    w(t)\left(\epsilon_{\phi}(\mathbf{z}_t; \mathbf{y}, t, \Delta \mathbf{p})-\epsilon\right)\frac{\partial \mathbf{z}}{\partial \theta},
\end{equation}
where $t$ is the timestep, $\mathbf{z}$ is the encoding of the rendered image, $\mathbf{z}_t$ is noisy latent at timestep $t$, $w(t)$ is a weighting function, $\epsilon_{\phi}$ is the noise predicted by the 2D diffusion prior, and $\Delta \mathbf{p}$ is the relative camera pose.


\vspace{-0.1in}
\subsection{Different Prior Knowledge Guidance}
\label{sec:Prior}
We introduce the proposed method by first analyzing the role of different priors in improving generation quality. Specifically, we address challenges related to initializing and optimizing Gaussian representations. Unlike reconstruction tasks, structure-from-motion (SfM) methods cannot be applied due to the only one input image. Additionally, optimization becomes challenging given only a frontal viewpoint, limiting the guidance available for Gaussian training.

\subsubsection{Geometry Prior} 
We assume that geometry prior provides the initialization information for the single input image.
A previous work \cite{radford2021learning} has shown that extending the geometric representation can achieve alignment between the geometrical data and the CLIP embedding space.
The geometry prior can be acquired through retrieving structurally analogous 3D shapes from an external dataset, following \cite{chen2024sculpt3d}, which chooses to incorporate external 3D prior information into text-to-3D generation by designing a shape supervision loss based on prior 3D shapes.
However, it is crucial to recognize the inherent distinction between these two generation tasks: while text-to-3D generation emphasizes high-level semantic alignment with text descriptions, image-to-3D generation requires that the frontal perspective of the generated object is exactly the same as the provided image in pixel level.
Directly applying 3D shape supervision loss may lead to conflicts between the input perspective and the generated perspective, so we adopt the retrieved 3D shapes to initialize Gaussian positions so as to ensure the coarse shape representation, which makes it easier to converge to the target shape after optimization. 

Specifically, we use the recently released OpenShape \cite{liu2023openshape} model, which scales up the 3D backbone to align with CLIP as our 3D retrieval module. The retrieved 3D shape provides coarse shape reference for the generation process.

\subsubsection{Perception Prior}

2D diffusion models have demonstrated powerful capabilities in image generation. Zero-1-to-3 \cite{liu2023zero} fine-tunes diffusion models to synthesize arbitrary views of an object given only a single frontal view, while SyncDreamer~\cite{liu2023syncdreamer} and MVDFusion~\cite{hu2024mvd} further extend this concept by modeling joint probability distributions of multiview images, enabling simultaneous multiview image generation.

Using SyncDreamer~\cite{liu2023syncdreamer}, we generate corresponding multiview images from the input image to guide the subsequent 3D Gaussian optimization. Despite the fact that the generation of additional views alleviates problems such as back-view blur to a certain extent, the resulting sparse and low-quality synthetic images are insufficient for reliable structure-from-motion (SfM) reconstruction to obtain accurate point clouds and poses, which plays a crucial role in the initialization of 3D Gaussians.

Following \cite{ke2024repurposing}, we consider to address this limitation by incorporating depth information. Specifically, we utilize Marigold to estimate rough depth maps from both the input and generated multiview images. We combine them to form a coarse point cloud to initialize the 3D Gaussian Splatting model, which provides a good solution for initialization issues.

\vspace{-0.1in}
\subsection{Integration of Different Gaussian branches}
\label{sec:consistency}

Similar to \cite{tang2023dreamgaussian}, we adopt a simple initialization strategy, randomly sampling Gaussian positions within a unit sphere. Although this approach lacks object-specific prior information, it remains generally effective for generating a broad variety of objects. Finally, we employ the geometry prior guidance, the perception prior guidance, and random Gaussian sampling strategy to initialize different Gaussian branches, respectively.

Each type of initialization brings unique advantages, motivating us to integrate beneficial aspects across different 3D Gaussian branches. 
We observe distinct behaviors among these Gaussian branches during iterative optimization, which are influenced by their initializations, as well as ambiguities in the 3D-to-2D projection. These factors, combined with the Gaussian densification process, lead to poor generation quality.
To overcome this, we introduce perceptual augmentation loss and stable SDS loss to enhance the multiview consistency of the model.



\subsubsection{Perceptual Augmentation Loss}
Specifically, we denote that the input image is $\tilde{\mathbf{I}}_{R}^0$ with a foreground mask $\tilde{\mathbf{I}}_{A}^0$, the generated $i$-th multiview image and its foreground mask\footnote{We utilize the Rembg library \cite{Gatisrembg2025} for isolating foreground objects to obtain the foreground mask.} are $\tilde{\mathbf{I}}_{R}^i$ and $\tilde{\mathbf{I}}_{A}^i$, where $i>0$. For each 3D Gaussian branch, we optimize the reference view image $\mathbf{I}_{R}^0$ and mask $\mathbf{I}_A^0$ to align with the input:
\begin{equation}
    \mathcal{L}_{\text{front}}=\lambda_1||\tilde{\mathbf{I}}_{R}^0-\mathbf{I}_{R}^0||_2^2+
    \lambda_2||\tilde{\mathbf{I}}_{A}^0-\mathbf{I}_{A}^0||_2^2,
  \label{eq:front_loss}
\end{equation}
where $\lambda_1$ and $\lambda_2$ are balancing parameters.

In addition to the mean-squared-error, we add a D-SSIM \cite{wang2004image} term to promote the structural similarity:
\begin{equation}
\begin{gathered}
    \mathcal{L}_{\text{aux}}=\lambda_3||\tilde{\mathbf{I}}_{R}^i-\mathbf{I}_{R}^i||_2^2+
    \lambda_4||\tilde{\mathbf{I}}_{A}^i-\mathbf{I}_{A}^i||_2^2+\lambda_5\mathcal{L}_{\text{D-SSIM}}(\tilde{\mathbf{I}}_{R}^i,\mathbf{I}_{R}^i),\\
    \text{for}\ \ i > 0,
\end{gathered}
  \label{eq:other view loss}
\end{equation}
where $\lambda_3$, $\lambda_4$ and $\lambda_5$ are balancing parameters.

\begin{figure}[t]
  \vspace{-0.15in}
  \centering
  \includegraphics[width=0.45\textwidth]{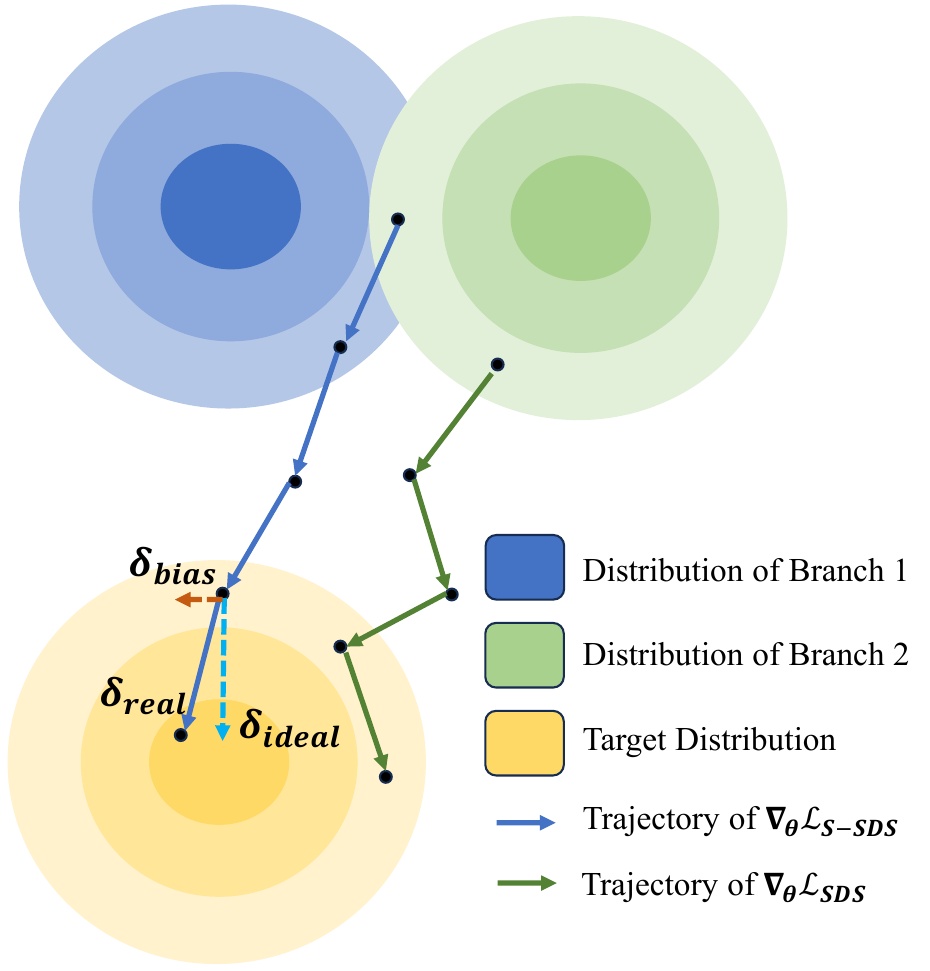}
  \caption{
  Conceptual figure of the proposed stable SDS between two Gaussian branches. The target distribution in the figure
 represents the conditional distribution of Gaussian parameters relative to the input image. Let $\delta_{\text{ideal}}=\delta_{\text{real}}+\delta_{\text{cond}}$ denote the ideal gradient update direction, while  $\delta_{\text{real}}=\delta_{\text{ideal}}+\delta_{\text{bias}}$ represents the actual update direction obtained in practice. Compared to $\nabla_{\theta}\mathcal{L}_{\text{SDS}}$, $\nabla_{\theta}\mathcal{L}_{\text{S-SDS}}$ further constrains the stochasticity of bias perturbations, thereby promoting more stable parameter updates that converge toward higher-density regions of the target distribution.}
  \label{fig:s-sds}
  \vspace{-0.2in}
\end{figure}

\subsubsection{Stable SDS Loss}
\label{C-SDS}
The SDS loss mentioned in Sec. \ref{sec:Preliminaries} introduces noise to the encoding of images rendered by 3D Gaussian Splatting from multiple perspectives and subsequently estimates the noise in the perturbed encoding using the input image as a conditional guide.
This method distills prior knowledge from a pretrained 2D diffusion model, incrementally updating the parameters of the 3D representation in the direction of the gradient, and produces images that exhibit a higher degree of fidelity aligned with the condition.

Tang et al. \cite{tang2023dreamgaussian} introduce SDS into the optimization process of 3D Gaussian Splatting. Nevertheless, as noted by the authors, the standard SDS may lead to oversaturated textures or blurry back-view textures.
Motivated by these findings, we analyze the underlying factors responsible for the problems as follows.

Ideally, the standard SDS gradient $\nabla_{\theta}\mathcal{L}_{\text{SDS}}$ can be decomposed into two key components: $\delta_{\text{cond}}$ and $\delta_{\text{real}}$. Here $\delta_{\text{cond}}$ guides the noisy latent $\mathbf{z}_t$ of the image $\mathbf{x}$ toward higher-density regions of conditional distribution $p(\mathbf{x}_t|\mathbf{y})$ (while the condition $\mathbf{y}$ is the input image). $\delta_{\text{real}}$ accounts for the discrepancy between the distributions of real-image and rendered-image.
In practical implementation, the noise added to the image encoding is sampled from a Gaussian prior, while the network predicts the noise distribution, introducing a residual bias $\delta_{\text{bias}}$. 
Consequently, even at convergence, the true SDS gradient effectively becomes:
\begin{equation}
    \nabla_{\theta}\mathcal{L}_{\text{SDS}}=\delta_{\text{cond}}+\delta_{\text{real}}+\delta_{\text{bias}}.
\end{equation}
We note that $\delta_{\text{cond}}$, $\delta_{\text{real}}$, and $\delta_{\text{bias}}$ are vectors. The undesired bias $\delta_{\text{bias}}$ interferes with the optimization process, leading to the localized oversaturation and blurring in the reconstructed 3D representation.

We next aim to decrease the influence of the residual bias $\delta_{\text{bias}}$ during optimization.
Our method leverages multiple Gaussian branches to provide complementary supervision throughout optimization. 
Once the two components $\delta_{\text{cond}}$ and $\delta_{\text{real}}$ have sufficiently converged, and the stochasticity introduced by bias $\delta_{\text{bias}}$ can lead the parameters of 3D Gaussians to drift arbitrarily, preventing the parameters from converging toward the high-density regions of the target distribution. 
Specifically, for each Gaussian branch, we leverage the other two branches to compute an averaged bias. The updated gradients are the weighted sum of its own $\delta_{\text{bias}}$ and the noise prediction adjusted by the other two reference branches.

Formally, after $l$ iterations,  we randomly sample a viewpoint and calculate its camera pose $\mathbf{p}$.
Let $\mathbf{x}^i$ be the image rendered by the $i$-th Gaussian branch under the camera pose $\mathbf{p}$, and let $\mathbf{z}_i$ be the corresponding latent encoding. We then define the stable SDS gradient for Gaussian branch $i$ as:

\begin{equation}
\begin{split}
    & \nabla_{\theta}\mathcal{L}_{\text{S-SDS}} = {}  \bigl[ w_1(t)(\epsilon_{\phi}(\mathbf{z}^{i}_{t};\mathbf{I}_R^0,t,\Delta \mathbf{p}) - \epsilon ) +\\
    & w_2(t) ( \epsilon_{\phi}(\mathbf{z}^{i}_{t};\mathbf{I}_R^0,t,\Delta \mathbf{p}) - \epsilon_{\phi}(\mathbf{z}^{j}_{t};\mathbf{I}_R^0,t,\Delta \mathbf{p}) ) +\\
    & w_3(t) ( \epsilon_{\phi}(\mathbf{z}^{i}_{t};\mathbf{I}_R^0,t,\Delta \mathbf{p}) - \epsilon_{\phi}(\mathbf{z}^{k}_{t};\mathbf{I}_R^0,t,\Delta \mathbf{p}) ) \bigr] \frac{\partial \mathbf{z}^i}{\partial \theta},
\end{split}
\label{eq:csds}
\end{equation}
where $j$, $k$ are the other two Gaussian branches. $\mathbf{z}_t^i$, $\mathbf{z}_t^j$ and $\mathbf{z}_t^k$ share the same sample noise $\epsilon$.

Eq. \ref{eq:csds} admits an interpretation through the lens of contrastive learning. Specifically, positive samples are critical in contrastive learning as they often share similar latent space. 
In our task, images rendered from different Gaussian branches under the same camera pose $\mathbf{p}$ are regarded as positive samples, which have the same conditions (the input image and the same camera pose), and after $l$ iterations, the images rendered from the different Gaussian branches are similar at the pixel level.
We then analyze the three components of different Gaussian branches $i$ and $j$. We utilize alphabetic superscripts to differentiate among Gaussian branches. 
For a camera pose $\mathbf{p}$, we have $\delta_{\text{cond}}^i \approx \delta_{\text{cond}}^j$ and $\delta_{\text{real}}^i \approx \delta_{\text{real}}^j$ after $l$ iterations. 
This derivation holds under the condition that both components reach convergence within $l$ iterative steps.
For bias $\delta_{\text{bias}}$, $\delta_{\text{bias}}^i \neq  \delta_{\text{bias}}^j$ because of its stochasticity. 
The term $( \epsilon_{\phi}(\mathbf{z}_t^i,...)-\epsilon_{\phi}(\mathbf{z}_t^j,...) )$ in Eq. \ref{eq:csds} can be approximated as $( \delta_{\text{bias}}^i-\delta_{\text{bias}}^j )$. 
The term $( \epsilon_{\phi}(\mathbf{z}_t^i,...)-\epsilon )$ can be approximated as $\delta_{\text{bias}}$.
So we approximate Eq. \ref{eq:csds} as:

\begin{equation}
\begin{split}
    \nabla_{\theta}\mathcal{L}_{\text{S-SDS}} = {}  &\bigl[ w_1(t)  \delta_{\text{bias}}^i  +
     w_2(t)  (\delta_{\text{bias}}^i - \delta_{\text{bias}}^j ) +\\
    & w_3(t)  (\delta_{\text{bias}}^i - \delta_{\text{bias}}^k ) \bigr] \frac{\partial \mathbf{z}^i}{\partial \theta}.
\end{split}
\label{eq:csds2}
\end{equation}

Compared to the standard SDS (the first term $\delta_{\text{bias}}$), the proposed stable SDS introduces two additional terms that constrain  $\delta_{\text{bias}}$ along a narrow directional range, whereas the standard SDS only regulates its magnitude. This directional restriction further reduces the stochasticity of $\delta_{\text{bias}}$, thus enhancing the stability of generation.  The conceptual illustration of the proposed stable SDS is shown in Fig. \ref{fig:s-sds}. 
 


\begin{figure}[t]
  \vspace{-0.1in}
  \centering
  \includegraphics[width=0.4\textwidth]{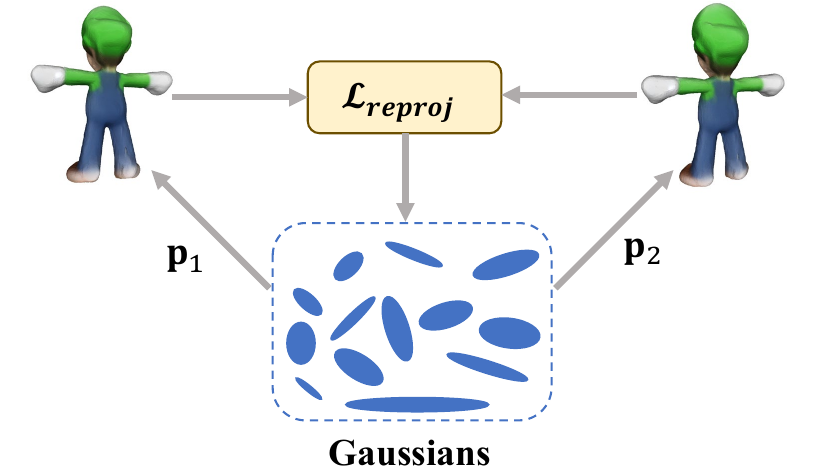}
  \caption{The process of calculating the reprojection loss. For a Gaussian branch, we sample a pair of views $\mathbf{p}_1$ and $\mathbf{p}_2$ with minimal viewpoint difference. We render images from the two views and calculate their pixel-level correspondence. Finally, we calculate the reprojection loss according to the correspondence.}
  \label{fig:reproj}
  \vspace{-0.1in}
\end{figure}

\subsubsection{Reprojection Loss for Depth Consistency}

After projecting 3D spatial points onto the image plane, a discrepancy often exists between the projected positions and the actual observed pixel locations in the image. We aim to exploit this discrepancy as a supervisory signal to refine the 3D Gaussian branches, thereby promoting local geometric consistency of the 3D representation.

To this end, we sample a pair of views $\mathbf{p}_1$ and $\mathbf{p}_2$ with minimal viewpoint difference, enabling us to approximate local correspondences between views and effectively guide the optimization process. Their camera extrinsic parameters are $\mathbf{M}_1 \in\mathbb{R}^{4\times4}$ and $\mathbf{M}_2 \in \mathbb{R}^{4\times4}$ from which we can get rotation and translation information for both views. For view 1, the rotation matrix is $\mathbf{R}_1=\mathbf{M}_{1,[0:3,0:3]}^{-1}$ and the translation vector is $\mathbf{T}_1=\mathbf{M}_{1,[0:3,3]}^{-1}$. $\mathbf{R}_2$ and $\mathbf{T}_2$ are in the same manner. With camera intrinsic matrix $\mathbf{K}\in \mathbb{R}^{3\times3}$, we need to calculate the corresponding position of each pixel of view 1 in view 2. For pixel $\mathbf{u}=(x,y,1)^\top \in \mathbb{R}^{3\times1}$ where $x$, $y$ represent the row and colume index, we first transform it to the world coordinate system:
\begin{equation}
    (w_x,w_y,w_z)^\top=\mathbf{R}_1^{-1}(\mathbf{K}^{-1}S\mathbf{u}-\mathbf{T}_1),
\end{equation}
where $S$ is the scaling factor and $w_x,w_y,w_z$ are the three components of the world coordinate system. Then we convert the point in the world coordinate system to another view:
\begin{equation}
    \mathbf{u'}=S^{-1}\mathbf{K}(\mathbf{R}_2(w_x,w_y,w_z)^\top+\mathbf{T}_2),
\end{equation}
assert $T_1=T_2=T$, then we obtain:
\begin{equation}
    \mathbf{u'}=\mathbf{K}\mathbf{R}_2\mathbf{R}_1^{-1}\mathbf{K}^{-1}\mathbf{u}+S^{-1}\mathbf{K}(\mathbf{1}-\mathbf{R}_2\mathbf{R}_1^{-1})\mathbf{T}.
    \label{eq:reproj}
\end{equation}
The scaling factor $S$ can be approximated as infinite, then  Eq. \ref{eq:reproj} can be approximated as:
\begin{equation}
    \mathbf{u'}=\mathbf{K}\mathbf{R}_2\mathbf{R}_1^{-1}\mathbf{K}^{-1}\mathbf{u}.
\end{equation}

Through the above deduction, we can determine the corresponding position $\mathbf{u'}$ in view 2 for the pixel $\mathbf{u}$ from view 1. We aim for the projected image to closely resemble the image rendered by the 3D Gaussian in view 2. Accordingly, we design the depth consistency loss as the $l_2$ norm between the estimated pixel values and the ground truth pixel values, formulated as:
\begin{equation}
    \mathcal{L}_{\text{proj}}=\frac{1}{|\text{D}_2|}\sum_{\mathbf{u}\in \text{D}_1}\lVert \text{D}_1(\mathbf{u})-\text{D}_2(\mathbf{u'})\rVert_2^2,
\end{equation}
where $\text{D}_1$ and $\text{D}_2$ are the depth images rendered by 3D Gaussians in view 1 and view 2, respectively, and $|\text{D}_2|$ means the total pixels in the depth image. The process of calculating the reprojection loss is shown in Fig. \ref{fig:reproj}.

\begin{figure*}[t]
  \centering
  \includegraphics[width=\textwidth]{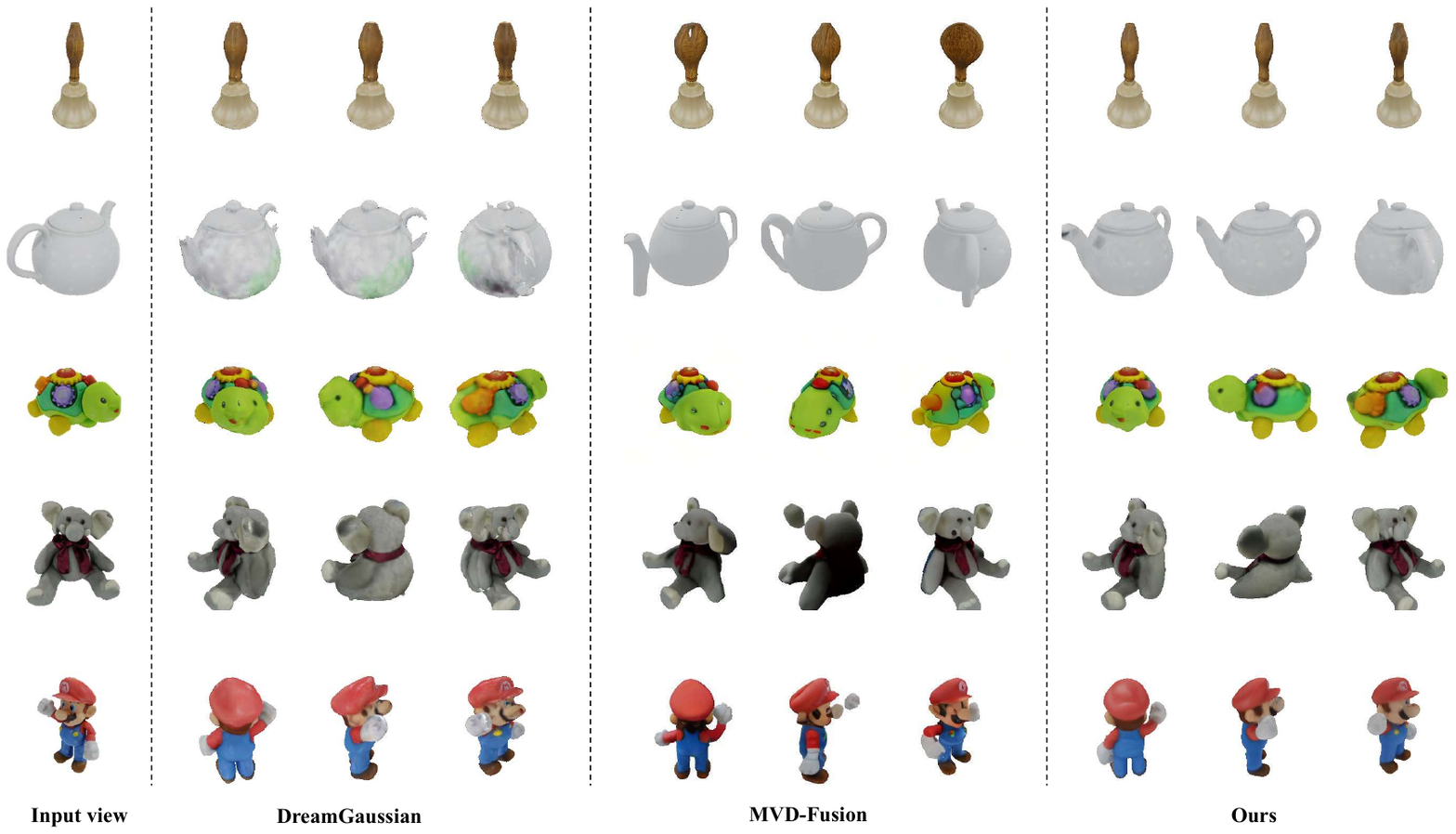}
  \vspace{-0.2in}
  \caption{Qualitative comparison with MVD-Fusion and DreamGaussian on synthesized multiview images.}
  \label{pic:nvs1}
  \vspace{-0.2in}
\end{figure*}

\vspace{-0.15in}
\subsection{Training Strategy}
\label{training}
Given an input image, we perform several pre-processing steps to initialize the Gaussian branches.

Firstly, we leverage a retrieval-based approach \cite{liu2023openshape} to identify the most similar 3D object from a reference dataset, which serves as the initialization for the first 3D Gaussian branch.  
Secondly, we utilize 2D priors \cite{liu2023syncdreamer} to synthesize multiple novel views of the input image. For each generated view, we apply a monocular depth estimation model \cite{ke2024repurposing} to infer depth maps, which are subsequently converted into point clouds. These point clouds are used to initialize the second 3D Gaussian branch.

When the iteration number is less than $l$ mentioned in Sec. \ref{C-SDS}), we jointly refine the three Gaussian branches using the following losses: a front-view alignment loss $\mathcal{L}_{\text{front}}$, an auxiliary multiview supervision loss $\mathcal{L}_{\text{aux}}$, a local projection consistency loss $\mathcal{L}_{\text{proj}}$, and the SDS loss  $\mathcal{L}_{\text{SDS}}$. The overall objective at this stage is:
\begin{equation}
    \mathcal{L}_1 = \mathcal{L}_{\text{front}} + \mathcal{L}_{\text{aux}} + \alpha \mathcal{L}_{\text{proj}} +  \mathcal{L}_{\text{SDS}},
\end{equation}
where $\alpha$ is a balancing parameter.

After $l$ iterations, we propose the stable SDS loss to further eliminate blurry and oversaturated textures. The final objective function becomes:
\begin{equation}
    \mathcal{L} = \mathcal{L}_1 + \beta \mathcal{L}_{\text{S-SDS}},
\end{equation}
where $\beta$ is a balancing parameter.

\vspace{-0.2in}
\section{Experiments}

We test the proposed method on the Google Scanned Object dataset (introduced in Sec. \ref{setup}) for the novel view synthesis task (in Sec. \ref{novel}) and single view reconstruction (in Sec. \ref{reconstruction}) without training any neural network. 
We show that our results achieve more accurate view synthesis as well as yield better 3D reconstruction compared to previous lightweight approaches.
We also present ablation studies (in Sec. \ref{ablation}) on crucial modules mentioned in Sec. \ref{sec:consistency}.

\vspace{-0.2in}
\subsection{Experimental Setup}
\label{setup}
\subsubsection{Evaluation dataset} 

We adopt the Google Scanned Object (GSO) \cite{downs2022google} dataset, which consists of high-quality scanned items from daily objects to animals, as the evaluation dataset. 
For each object, we render 16 views evenly spaced in azimuth from an elevation of 30 degrees, and choose the frontal one of them as the input image. For quantitative results, we randomly choose 30 objects to
 compute the metrics.
 
\vspace{-0.2in}
\subsubsection{Implementation Details}

We train 1200 steps for each object and set $l=600$. 
Sec. \ref{training} clarified the initialization of first and second Gaussian branches. For the third branch, we sample points inside a sphere of radius 0.5. The number of points is 10k for the three initializations. For the first branch, we introduce mesh sampling and farthest point sampling (FPS) to downsample the 3D object to 10k points. For the second branch, we introduce FPS to downsample to 10k points.

All parameters of the loss function except $\mathcal{L}_{\text{SDS}}$ and $\mathcal{L}_{\text{S-SDS}}$ are increased linearly from 0 to a maximum value. We show the maximum value of related parameters in Table \ref{tab:parameter}.
\begin{table}[!btp]
	\caption{The maximum value of related parameters. }
    \vspace{-0.1in}
	\begin{center}
		{
			\begin{tabular}{cccccccc}
				\toprule
				 $\lambda_1$ & $\lambda_2$ & $\lambda_3$ &$\lambda_4$ & $\lambda_5$ & $\alpha$ \\
				\midrule
10000 &  1000 &  4000 &  200 & 100 & 50 \\
				\bottomrule
			\end{tabular} 
		}
	\end{center}
	\label{tab:parameter}
  \vspace{-0.2in}
\end{table}
For $\mathcal{L}_{\text{SDS}}$ and $\mathcal{L}_{\text{S-SDS}}$, $t\sim \mathcal{U}(0.02,0.98)$ is a randomly sampled timestep, $p$ is a randomly sampled camera pose orbiting the object center, $\epsilon \sim \mathcal{N}(0,1)$ is a random Gaussian noise, $w(t) = \sigma^2_t$ is a weighting function from DDPM \cite{ho2020denoising}, $w_2(t)=w_3(t)=\frac{1}{2}w_1(t)$, and we set $\beta=1.5$.

In training our 3D Gaussian Splatting model, we assign distinct parameter values tailored to each attribute. The learning rate for position decays from $2 \times 10^{-4}$ to $1 \times 10^{-6}$ over 1200 steps, while those for features, opacity, scaling, and rotation are set to $0.0025$, $0.05$, $0.0025$, and $1 \times 10^{-3}$, respectively.
Densification is applied to Gaussians whose accumulated gradient exceeds 0.1 or whose maximum scaling falls below 0.05. We prune Gaussians with opacity lower than $0.025$ or maximum scaling greater than $0.1$.
For the mesh refinement module, we follow the same approach as DreamGaussian \cite{tang2023dreamgaussian}. 
All experiments are conducted on a single RTX 3090 GPU.

\subsubsection{Baseline}

For the novel view synthesis task, we adopt Zero123 \cite{liu2023zero}, RealFusion \cite{melas2023realfusion}, SyncDreamer \cite{liu2023syncdreamer}, DreamGaussian \cite{tang2023dreamgaussian} and MVD-Fusion \cite{hu2024mvd} as baseline methods.
Given an input image of an object and target view parameters, Zero123 can estimate the image of the target view. 
RealFusion is based on Stable Diffusion \cite{rombach2022high} and the SDS loss for single-view reconstruction.
SyncDreamer and MVDFusion can generate multiview images simultaneously based on a fine-turned 2D diffusion model.
DreamGaussian adopts 3D Gaussian Splatting as 3D representation and only uses an input image and the SDS to optimize the object without training additional models.

For single view reconstruction, we adopt Zero123 \cite{liu2023zero}, RealFusion \cite{melas2023realfusion}, Magic123 \cite{qian2023magic123}, One-2-3-45 \cite{liu2023one}, Point-E \cite{nichol2022point}, Shap-E \cite{jun2023shap}, DreamGaussian \cite{tang2023dreamgaussian}, SyncDreamer \cite{liu2023syncdreamer} and MVDFusion \cite{hu2024mvd} as baseline methods.
Magic123 combines Zero123 with RealFusion to further improve the quality of the reconstruction. 
One-2-3-45 directly regresses the sign distance function from the output images of Zero123.
Point-E and Shap-E are 3D generative models trained on a large internal OpenAI 3D dataset, both of which are able to convert a single-view image into a point cloud or a shape encoded in an MLP. For Point-E, we convert the generated point clouds to SDFs for shape reconstruction.
We should note that the diffusion-based methods require additional methods to render or distill objectives. For instance, Zero-1-to-3 requires an additional SDS distillation to generate 3D shapes whereas SyncDreamer relies on training Neus or NeRF.

\begin{figure*}[t]
  \centering
  \includegraphics[width=\textwidth]{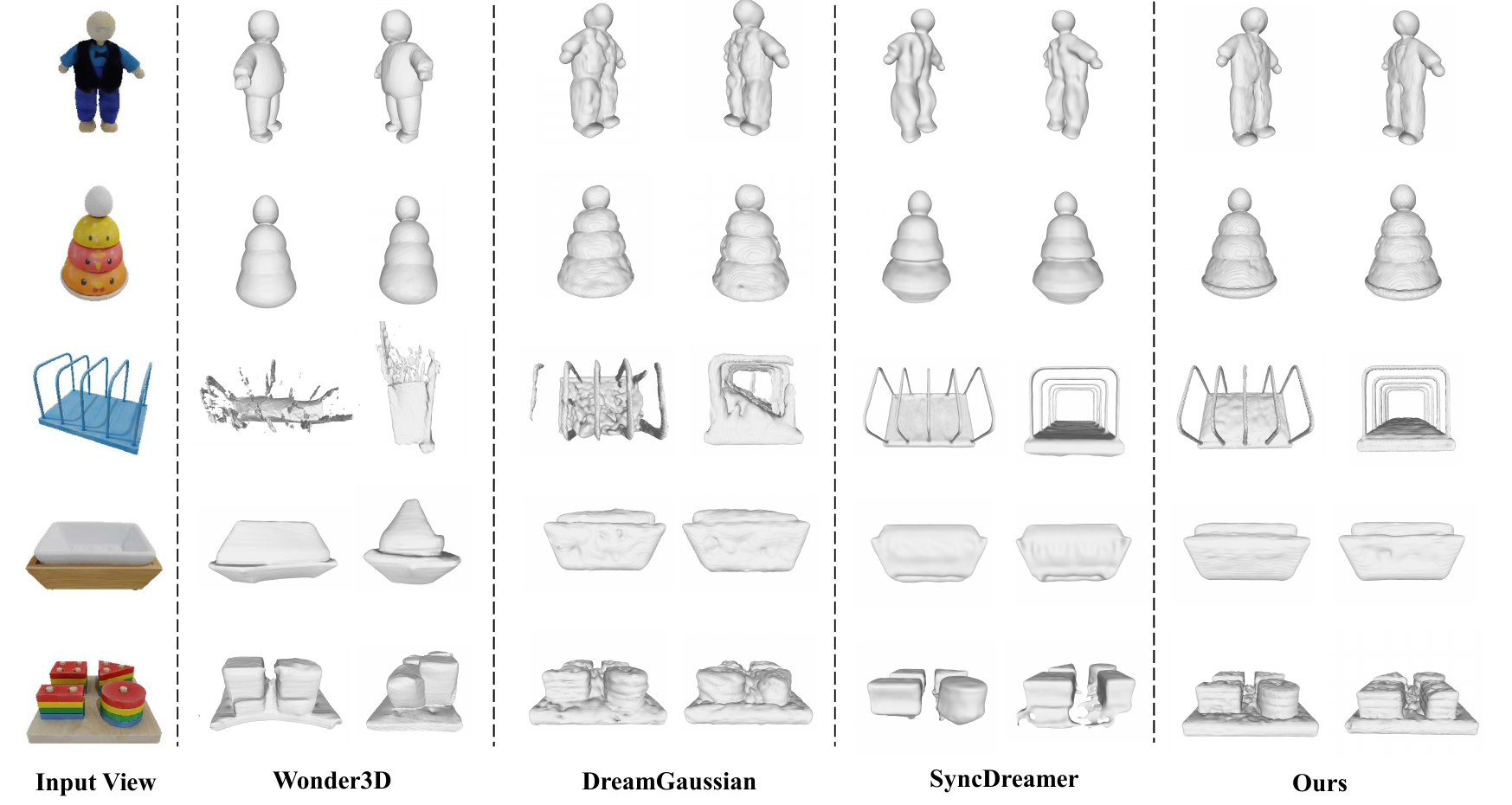}
  \vspace{-0.2in}
  \caption{Qualitative comparison of reconstruction from single view images with different methods.}
  \label{pic:recon1}
  \vspace{-0.2in}
\end{figure*}

\subsubsection{Metrics}

For novel view synthesis, we adopt commonly used metrics: PSNR, SSIM \cite{wang2004image}, and LPIPS \cite{zhang2018unreasonable}.
For single view reconstruction, we report the commonly used Chamfer Distance (CD) and Volume IoU between ground-truth shapes and reconstructed shapes.

\vspace{-0.2in}
\subsection{Novel View Synthesis}
\label{novel}
For this task, the quantitative results are shown in Table \ref{tab:nvs} and the qualitative results are shown in Fig. \ref{pic:nvs1}.
RealFusion \cite{melas2023realfusion} applies a NeRF model
 to distill the stable diffusion model, but is unable to produce visually plausible images.
Zero123 \cite{liu2023zero} produces visually reasonable images but suffers from inconsistency and instability among the generated images.
SyncDreamer \cite{liu2023syncdreamer} and MVD-Fusion \cite{hu2024mvd} are limited to specific perspectives. Meanwhiles, MVD-Fusion is prone to perspective inaccuracy and multiple-perspective inconsistency, as shown in the third column of Fig. \ref{pic:nvs1}.
DreamGaussian \cite{tang2023dreamgaussian} faces oversaturated textures or blurry back-view textures as shown in the second and fifth rows of Fig. \ref{pic:nvs1}.
Our method utilizes geometry and perception prior knowledge to supervise different Gaussian branches, achieving competitive generation quality.

\begin{table}[!btp]	
\caption{
The quantitative comparison in novel view synthesis. We report PSNR, SSIM, LPIPS on the GSO \cite{downs2022google} dataset. }
\vspace{-0.15in}
	\begin{center}
		{
			\begin{tabular}{ccccc}
				\toprule
				 Method &  Year & PSNR $\uparrow$ &   SSIM $\uparrow$ & LPIPS $ \downarrow$\\
				\midrule
 RealFusion \cite{melas2023realfusion} & 2023 & 15.26 & 0.722 & 0.283 \\
 Zero123 \cite{liu2023zero} & 2023 & 18.93 &  0.779 &  0.166 \\
 Wonder3D \cite{long2024wonder3d} & 2024 & 16.23 & 0.797 & 0.217 \\
 SyncDreamer \cite{liu2023syncdreamer} & 2024 & 20.05 &  0.798 &  0.146 \\
 MVD-Fusion \cite{hu2024mvd} & 2024 & 19.99 & 0.830 &  0.159 \\
 DreamGaussian \cite{tang2023dreamgaussian} & 2024 & 18.45 &  0.808 &  0.189 \\
				\midrule
 Ours & 2025 &  \textbf{21.24} &  \textbf{0.856} &  \textbf{0.132} \\
				\bottomrule
			\end{tabular} 
		}
	\end{center}
	\label{tab:nvs}
\vspace{-0.2in}
\end{table}

\begin{table}[!btp]	
\caption{The quantitative comparison in 3D reconstruction.  We report Chamfer Distance and Volume IoU on the GSO dataset.}
\vspace{-0.15in}
	\begin{center}
		{
			\begin{tabular}{cccc}
				\toprule
				 Method &  Year & Chamfer Distance$\downarrow$ &   Volume IoU $\uparrow$ \\
				\midrule
 Realfusion \cite{melas2023realfusion} & 2023 & 0.0819 & 0.2741 \\
 Magic123 \cite{qian2023magic123} & 2023 & 0.0516 &  0.4528 \\
 One-2-3-45 \cite{liu2023one} & 2023 & 0.0629 &  0.4086  \\
 Point-E \cite{nichol2022point} & 2022 & 0.0426 & 0.2875  \\
 Shape-E \cite{jun2023shap} & 2023 & 0.0436 &  0.3584  \\
 Zero123 \cite{liu2023zero} & 2023 & 0.0339 &  0.5035  \\
 Wonder3D \cite{long2024wonder3d} & 2024 & 0.0408 & 0.5171 \\
 SyncDreamer \cite{liu2023syncdreamer} & 2024 & 0.0261 & 0.5421 \\
 MVD-Fusion \cite{hu2024mvd} & 2024 & 0.0310 & - \\
 DreamGaussian \cite{tang2023dreamgaussian} & 2024 & 0.0327 & 0.5828 \\
				\midrule
 Ours & 2025 & \textbf{0.0186} & \textbf{0.6167} \\
				\bottomrule
			\end{tabular} 
		}
	\end{center}
\vspace{-0.2in}

	\label{tab:recon}
\end{table}

\begin{table}[!btp]	
\vspace{-0.15in}
\caption{Training and inference consumption comparison in 3D reconstruction. The experiments on inference time and inference consumption are conducted on a single NVIDIA 3090 GPU.}
\vspace{-0.15in}
	\begin{center}
		{
			\begin{tabular}{ccccc}
				\toprule
				 Method &  \makecell{Training \\ Time} &  \makecell{Training \\ Consum.} & \makecell{Inference \\ Time} & \makecell{Inference \\ Consum.} \\
				\midrule
                MVD-Fusion \cite{hu2024mvd} & 7 days & \makecell{4 A100\\GPUs} & 2 min & 20 GB\\
				\midrule
                Zero123 \cite{liu2023zero} & 7 days& \makecell{8 A100\\GPUs} & 15 min & 20 GB \\ 
                SyncDreamer \cite{liu2023syncdreamer} & 4 days & \makecell{8 A100\\GPUs} & 20 min & 20 GB \\ 
                Wonder3D \cite{long2024wonder3d} & 3 days & \makecell{8 A800\\GPUs} & 5 min & 24 GB \\
                DreamGaussian \cite{tang2023dreamgaussian} & - & - & 2 min & 10 GB \\
				\midrule
                Ours & - & - & 10 min &  10 GB \\
				\bottomrule
			\end{tabular} 
		}
	\end{center}
\vspace{-0.15in}
\label{tab:consume}
\end{table}

\vspace{-0.2in}
\subsection{Single-view Reconstruction}
\label{reconstruction}
For this task, we show the quantitative results in Table \ref{tab:recon} and the qualitative comparison in Fig \ref{pic:recon1}.
Magic123 \cite{lin2023magic3d} heavily relies on the estimated depth values of the input view, leading to incorrect results and instability. 
Point-E \cite{nichol2022point} and Shap-E \cite{jun2023shap} tend to produce incomplete meshes.
One-2-3-45 \cite{liu2023one} reconstructs meshes from the outputs of Zero123, but the reconstruction suffers from loss of details due to image quality constraints.
SyncDreamer \cite{liu2023syncdreamer} uses vanilla NeuS \cite{wang2021neus} for reconstruction. The method is incapable of direct 3D mesh generation. Even with other methods, its texture quality still cannot reach the ideal standard.
MVD-Fusion \cite{hu2024mvd} can directly obtain the point cloud but cannot generate meshes.
DreamGaussian \cite{tang2023dreamgaussian} generates imcomplete 3D geometric structure in some cases, as shown in the third row of Fig. \ref{pic:recon1}. 
Wonder3D \cite{long2024wonder3d} produces preliminary geometric approximations of objects, yet exhibits limitations in fine-grained detail reproduction and occasional inaccuracies in 3D structural reconstruction.
Comparative evaluations demonstrate that our approach significantly improves reconstruction fidelity while simultaneously maintaining surface continuity and preserving high-frequency geometric features.

Furthermore, we compare the consumption of different methods in the single-view reconstruction task, as shown in Table \ref{tab:consume}.
MVD-Fusion, Zero123, SyncDreamer and Wonder3D are based on fine-turning diffusion models to improve generation consistency, which leads to a considerable amount of time and GPU consumption.
Meanwhile, although MVD-Fusion exhibits rapid inference, it can only generate point clouds instead of meshes without other tools.
Although DreamGaussian achieves notable improvements in computational efficiency and resource utilization over prior approaches, limitations persist in the quality of its output textures and geometric structures, suggesting avenues for future enhancement.
Our method achieves high-quality generation without the need to train additional neural networks, while also maintaining low computational cost during inference.

\begin{figure*}[t]
  \centering
  \includegraphics[width=\textwidth]{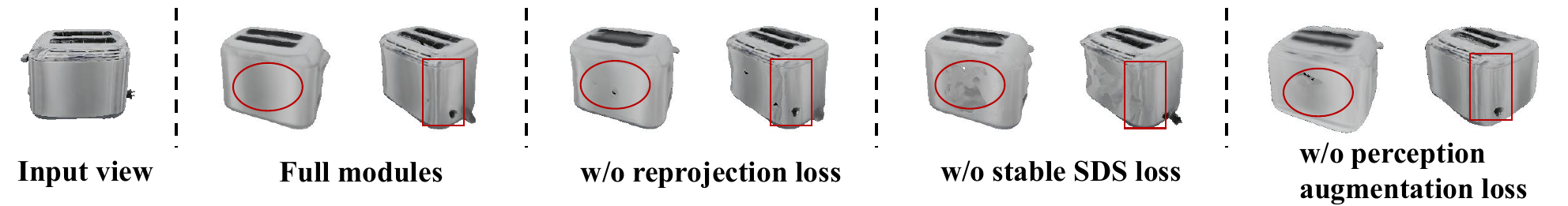}
  \vspace{-0.2in}
  \caption{Ablation study on the different module on novel view synthesis task.}
  \label{pic:ablation}
  \vspace{-0.2in}
\end{figure*}

\vspace{-0.2in}
\subsection{Ablation Study}
\label{ablation}

\subsubsection{Different Branches}

\begin{table}[!btp]	
\caption{The impact of different branches on novel view synthesis task. }
\vspace{-0.15in}
	\begin{center}
		{
			\begin{tabular}{cccc}
				\toprule
				 Branch &  PSNR $\uparrow$ &   SSIM $\uparrow$ & LPIPS $ \downarrow$\\
				\midrule
 3D Shape Retrieval &  20.74 &  0.846 &  0.144 \\
 Depth Estimation &  20.75 & 0.849 &  0.142 \\
 Noise &  20.63 &  0.847 &  0.144 \\
 All &  \textbf{21.24} &  \textbf{0.856} &  \textbf{0.132} \\
				\bottomrule
			\end{tabular} 
		}
	\end{center}
	\label{tab:ab_branch}
    \vspace{-0.2in}
\end{table}
To show how different Gaussian branches affect the result and the improvement of interactions in branches, we conduct experiments with only one of the three Gaussian branches without interactions ({\it e.g.}, the stable SDS). As we see in Table \ref{tab:ab_branch}, only one branch shows decrease in three indicators, demonstrating the necessity of the interaction of different branches. 

\subsubsection{Different Modules}

To show how different modules mentioned in Sec. \ref{sec:consistency} contribute to the generation results, we discard perceptual augmentation loss, stable SDS loss, and reprojection loss,  respectively. 
As we see in Table~\ref{tab:ab_module}, the absence of any module will lead to a deterioration of quantitative results.
We also display the novel view synthesis results based on the absence of the proposed modules. As in Fig.~\ref{pic:ablation}, the absence of reprojection loss leads to local noise, while the absence of stable SDS loss and perception augmentation loss leads to blurry or oversaturated textures.
The results with full modules exhibit reduced blurriness and improved accuracy.

\begin{table}[!btp]	
\caption{The impact of different module on novel view synthesis task. }
\vspace{-0.15in}
	\begin{center}
		{
			\begin{tabular}{cccc}
				\toprule
				 Module &  PSNR $\uparrow$ &   SSIM $\uparrow$ & LPIPS $ \downarrow$\\
				\midrule
 w/o Perception Augmentation Loss& 20.62 & 0.844 & 0.147\\
 w/o Co-SDS Loss&  20.68 &  0.847 &  0.140 \\
 w/o Reprojection Loss &  20.95 & 0.851 &  0.136 \\
 All &  \textbf{21.24} &  \textbf{0.856} &  \textbf{0.132} \\
				\bottomrule
			\end{tabular} 
		}
	\end{center}

	\label{tab:ab_module}
\vspace{-0.2in}
\end{table}


\vspace{-0.2in}
\section{Conclusion}
We propose a novel method for reconstructing detailed 3D objects from a single image.
Our method employs Gaussian splatting, guided by geometry and perception priors, to address the challenges of high memory requirements.
To ensure effective knowledge transfer, we introduce a stable SDS loss for fine-grained prior distillation. 
Furthermore, we propose a reprojection-based strategy to reinforce geometric consistency in depth estimation. 
Experimental results demonstrate that our approach achieves superior generation fidelity, effectively improving the multiview consistency and geometric detail.
\bibliographystyle{ieeetr}
\vspace{-0.2in}
\bibliography{reference.bib}  

\end{document}